\documentclass[11pt,a4paper]{article}
\usepackage[hyperref]{ranlp2021}
\usepackage{times}
\usepackage{latexsym}

\usepackage{microtype}

\usepackage{rotating}

\usepackage{makecell}
\usepackage{booktabs}
\usepackage{multirow}
\usepackage{amsmath}
\usepackage{tikz,pgfplots}
\usetikzlibrary{arrows,shapes.misc,chains,scopes}
\usetikzlibrary{decorations.pathreplacing}
\usepackage{graphicx}

\aclfinalcopy 

\title{Naturalness Evaluation of Natural Language Generation in Task-oriented Dialogues using BERT}

\author{Ye Liu$^1$, Wolfgang Maier$^2$, Wolfgang Minker$^3$ \and Stefan Ultes$^4$ \\
 $^{1,2,4}$Mercedes-Benz AG, Sindelfingen, Germany \\
 $^3$Ulm University, Ulm, Germany \\
 \texttt{ $^1$ye.y.liu@daimler.com} \\
 \texttt{ $^{2,4}$\{wolfgang.mw.maier,stefan.ultes\}@daimler.com}\\
 \texttt{ $^3$wolfgang.minker@uni-ulm.de}}

\date{}

\begin{document}
\maketitle
\begin{abstract}
This paper presents an automatic method to evaluate the naturalness of natural language generation in dialogue systems. While this task was previously rendered through expensive and time-consuming human labor, we present this novel task of automatic naturalness evaluation of generated language. By fine-tuning the BERT model, our proposed naturalness evaluation method shows robust results and outperforms the baselines: support vector machines, bi-directional LSTMs, and BLEURT. In addition, the training speed and evaluation performance of naturalness model are improved by transfer learning from quality and informativeness linguistic knowledge.
\end{abstract}

\section{Introduction}
\label{sec: introduction}
With the increasing popularity of virtual assistants such as Alexa or Siri, users have a higher demand for conducting natural conversations. They would like to chat with these assistants more naturally---maybe even like talking to a real human being. One of the key questions arising from this, though, is how to measure the naturalness of the generated language. In the past, native speakers judged the quality of the generated language by answering questions like ``is this utterance natural?'' or ``could it have been produced by a native speaker?'' to rate the naturalness~\citep{novikova2016crowd}. However, this approach heavily depends on manual effort and is impractical for broader use. On the other hand, the widely used automatic metrics for evaluating language generation, like BLEU~\citep{papineni2002bleu} and METEOR~\citep{banerjee2005meteor}, rely on word overlap mechanism and compare the generated sentence to one or more human-created reference sentences~\citep{stent2005evaluating}, but cannot directly reflect naturalness information.

\begin{figure}[t]
\footnotesize
\begin{center}
\begin{tabular}{cccccccc}

\multicolumn{8}{c}{Sentence: X is a moderately priced restaurant in X.} 
\\
\\
\makecell{very \\ unnatural} & $\bigcirc$  &  $\bigcirc$ & $\bigcirc$ & $\bigcirc$ & $\bigcirc$ & $\bigcirc$ & \makecell{very \\ natural} 

\end{tabular}
\end{center}
\caption{\label{tab:6-point Likert scale for naturalness.} Example of 6-point Likert scale for evaluating the naturalness of a sentence.}
\end{figure}

Likert scale ratings are generally used in human evaluation. The Figure~\ref{tab:6-point Likert scale for naturalness.} shows an example of 6-point Likert scale rating on naturalness of a sentence. In general, the human evaluation items (naturalness, coherence, etc.) are hard explainable objectively and native speakers purely rely on their underlying criteria to judge the performance of generated language without any specified rules. To alleviate the manual effort previously needed, the goal of this paper is to render the task of assessing the naturalness of generated language as an automatic reference-based classification problem having each Likert scale rating representing one class. To our knowledge, we are the first to propose this task and present an approach evaluating naturalness of generation through fine-tuning a pre-trained Language Model (LM).

The contribution of this paper is two-fold: a pre-trained BERT model is fine-tuned to estimate the naturalness of a generated sentence based on a reference sentence. Three baselines are proposed for robust performance comparison: a support vector machine (SVM) baseline using bag-of-words (BoW) vectors for input representation, a bi-directional Long Short Term Memory (LSTM) neural network, and the pre-trained BLEURT model~\citep{sellam2020bleurt}. Second, based on the positive correlation between naturalness and other annotated information: quality and informativeness of the generated sentence, the proposed method is extended to leverage this additional information through transfer learning. And the learning speed and estimation performance of naturalness evaluation model is significantly increased. 

The remainder of this paper is structured as follows: Section~\ref{sec: related work} introduces the related works. In Section~\ref{sec: Naturalness Estimation Using BERT}, our proposed approach for BERT-based naturalness estimation is presented. In Section~\ref{sec: experiment setup}, the experiment setups are covered, which include fine-tuning BERT, comparison against the baselines, and transfer learning. Section~\ref{sec: Results and Discussion} describes the experiments results. The last Section~\ref{sec: Conclusion and Future Work} draws conclusions and outlines future work.

\section{Related Work}
\label{sec: related work}
With the development of Natural Language Generation (NLG) applications and their benchmark datasets, evaluation of NLG systems has become increasingly important. Generally, multiple automatic metrics are used in parallel to evaluate the performance of language generation, such as BLEU~\citep{papineni2002bleu}, METEOR~\citep{banerjee2005meteor} or ROUGE~\citep{lin2004rouge}. However, \citet{chaganty2018price} demonstrated that the existing automatic metrics have poor instance-level correlation with mean human judgment and that they assign bad scores to many good quality responses.

Since automatic metrics still fall short of replicating human decisions~\citep{krahmer2010empirical,reiter2018structured}, many NLG papers include some form of human evaluation. For example, \citet{hashimoto2019unifying} report that 20 out of 26 generation papers published at ACL2018 presented human evaluation results for showing their robust performance comparison. ~\citet{celikyilmaz2020evaluation} and~\citet{gatt2018survey} also highlighted that human evaluation is commonly viewed as the best reliable way to evaluate NLG systems, but come with many issues, such as costly and time consuming and human judgement bias. And more importantly, the evaluation results from human efforts are not always repeatable~\citep{belz2006comparing}. \citet{dusek2017referenceless} previously attempted to predict quality ratings of generated language by using Recurrent Neural Network (RNN) with the help of the Meaning Representations (MRs) and showed promising performance. However, our work is more focused on naturalness evaluation and based on the gold reference. In order to relieve the human labor, we propose a reference-based method for naturalness evaluation by utilizing neural network to learn the complicated linguistic relationship from sentences. And this work can be easily extended to other human evaluation criteria (coherence, quality, etc..), if the corresponding human evaluation data is available.

In recent years, with huge success of pre-trained LMs~\citep{devlin2019bert, radford2018improving}, many machine learned metrics for evaluating generation are proposed. Especially the pre-trained BERT (Bidirectional Encoder Representations from Transformers)~\citep{devlin2019bert} shows its superiority in this field.~\citet{shimanaka2019machine} presented automatic machine translation evaluation by using BERT and achieved the best performance in segment-level metrics tasks on the WMT17 dataset for all to-English language pairs.~\citet{zhang2019bertscore} proposed an automatic evaluation metric for text generation based on pre-trained BERT contextual embeddings: BERTScore. BERTScore computes the similarity of two sentences as a sum of cosine similarities between their tokens’ embeddings. And~\citet{zhang2019bertscore} showed BERTScore correlates better with human judgments and provides stronger model selection performance than existing metrics.~\citet{sellam2020bleurt} presented BLEURT, which continually pre-trained BERT on synthetic data and then fine-tuned on task-specific ratings. And~\citet{sellam2020bleurt} demonstrated BLEURT can model human assessment with superior accuracy. Given the superiority of BERT, we also apply the pre-trained BERT for our proposed naturalness evaluation on generated language.

\section{Naturalness Evaluation Using BERT}
\label{sec: Naturalness Estimation Using BERT}
The task of estimation the naturalness of a generated sentence is framed as a classification task. Two sentences are used as input: the candidate sentence to be scored and a reference sentence. The naturalness score is derived through fine-tuning a pre-trained BERT~\citep{devlin2019bert} model. The main architecture of BERT uses the encoder of a Transformer~\citep{vaswani2017attention}, which is an advanced encoder-decoder architecture leveraging the attention mechanism. Considering that the input may be sentence pairs in several tasks, technical innovation Next Sentence Prediction (NSP) helps BERT to learn the relationship between sentence pairs by receiving pairs of sentences as input and separating them with [SEP] token to learn predicting if the second sentence is the subsequent sentence in the original document. To do so, a [CLS] token is added at the beginning for every input to learn the meaning of the entire input. Exactly because of these specific characters, eleven NLP tasks in~\citet{devlin2019bert} obtained new state-of-the-art results by fine-tuning BERT. 

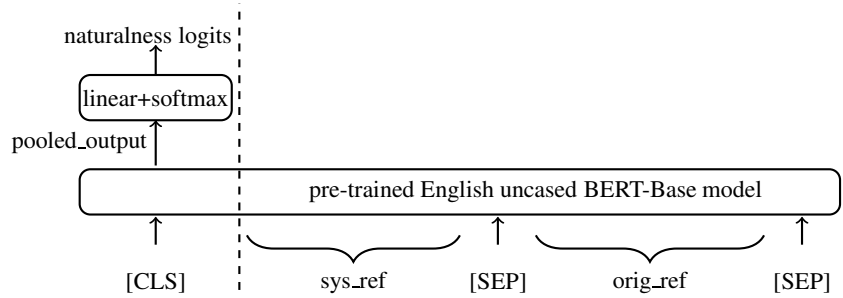
\begin{figure*}
\centering
\begin{tikzpicture}
\footnotesize
\draw[rounded corners,thick]   (0,0) rectangle (10,0.6);
\node at (6,0.3) {pre-trained English uncased BERT-Base model};

\draw[thick,->] (1,0.65) -- (1,1.25);
\node at (0,0.95) {pooled\_output};

\draw[rounded corners,thick]   (0,1.25) rectangle (2,1.85);
\node at (1,1.55) {linear+softmax};

\draw[thick,->] (1,1.85) -- (1,2.25);
\node at (0.9,2.35) {naturalness logits};

\draw[decorate,decoration={brace,amplitude=10pt,raise=5pt,mirror},yshift=-0.5cm,thick] (2.2,0.4) -- (5,0.4) node [black,midway,yshift=-0.8cm] {sys\_ref};
\draw[decorate,decoration={brace,amplitude=10pt,raise=5pt,mirror},yshift=-0.5cm,thick] (6,0.4) -- (9,0.4) node [black,midway,yshift=-0.8cm] {orig\_ref};
\draw[thick,->] (1,-0.4) -- (1,-0.03);
\node at (1,-0.9) {[CLS]};
\draw[thick,->] (5.5,-0.4) -- (5.5,-0.03);
\node at (5.5,-0.9) {[SEP]};
\draw[thick,->] (9.5,-0.4) -- (9.5,-0.03);
\node at (9.5,-0.9) {[SEP]};

\draw[thick,dashed] (2.1,-1) -- (2.1,2.8);

\end{tikzpicture}

\caption{Fine-tuning BERT architecture for naturalness estimation}
\label{fig:Fine-tuning BERT architecture}
\end{figure*}

Figure~\ref{fig:Fine-tuning BERT architecture} shows the fine-tuning structure for naturalness estimation with a candidate sentence (sys\_ref) and a reference sentence (orig\_ref) as input. Example for a sys\_ref and orig\_ref is shown in Table~\ref{tab: dataset example}. In accordance to the NSP task, representing both sentences on the input side is achieved by separating them with the [SEP] token. As can be seen in Figure~\ref{fig:Fine-tuning BERT architecture}, an additional [CLS] token is inserted at the beginning of the first sentence. The final output of the [CLS] token, which is called pooled\_output, forms a representation of the entire input. Then a linear layer with softmax activation is added to the top of pooled\_output to predict the probability of sentence-level naturalness label. To our knowledge, we are the first to fine-tune BERT to learn the abstract naturalness linguistic information and demonstrate the robust performance of this method. 


\begin{table}
\footnotesize
\begin{center}
\begin{tabular}{rlcccc}
\toprule
     & & \begin{sideways}judge\end{sideways}  &  \begin{sideways}\textbf{naturalness}\end{sideways} & \begin{sideways}quality\end{sideways} & \begin{sideways}informativeness\end{sideways} \\
    
   \cmidrule{3-6}
    sys\_ref:& \textit{zuni cafe, is expensive. } &  1 & \textbf{6} & {5} & {6}\\
   orig\_ref: & \multirow{2}{*}{\shortstack[l]{\textit{how about zuni cafe,}\\ \textit{an expensive one?}}} &  2 & \textbf{5} & {4} & {6}\\
    & &  3 & \textbf{6} & {4} & {5}\\
   
   \bottomrule
\end{tabular}
\end{center}
\caption{\label{tab: dataset example} Example of the pre-processed data set. }
\end{table}

\section{Experimental Setup}
\label{sec: experiment setup}
In this section, the experiment procedure including the pre-processing of the used data is introduced. First, BERT is fine-tuned for naturalness estimation and then compared it with the baselines. Furthermore, external knowledge is added demonstrating the impact on naturalness estimation through transfer learning.

\subsection{Dataset Preprocessing}
\label{subs: dataset collection}
The dataset\footnote{https://researchportal.hw.ac.uk/en/datasets/human-ratings-of-natural-language-generation-outputs}~\citep{novikova2017we} comprises textual dialog responses produced by three data-driven NLG systems over three different domains. The three NLG systems are respectively {\bf RNNLG}\footnote{https://github.com/shawnwun/RNNLG}, {\bf TGen}\footnote{https://github.com/UFAL-DSG/tgen} and {\bf LOLS}\footnote{https://github.com/glampouras/JLOLS}. The applied domains are {\bf SF Hotel} and {\bf SF Restaurant}~\citep{wen2015semantically}, which provide information about hotels and restaurants in San Francisco, and {\bf BAGEL}~\citep{mairesse2010phrase} that provides information about restaurants in Cambridge. Human annotations on naturalness, quality, and informativeness were collected for each NLG-produced text on a 6-point Likert scale by asking the annotator ``could the utterance have been produced by a native speaker?'', ``How do you judge the overall quality of the utterance in terms of its grammatical correctness and fluency?'', ``Does the utterance provide all the useful information from the meaning representation?'', respectively.

Table~\ref{tab: dataset example} shows an annotation example of the data. The sentence pair input comprises sys\_ref and orig\_ref. The sys\_ref presents the output of each of the above-mentioned three NLG systems, one at a time, while orig\_ref denotes human written references from the original data. The judges (1, 2, 3) represent the three human raters. The naturalness score is our target label. In addition to the naturalness score, the informativeness score and the quality score are utilised in transfer learning experiments to improve naturalness estimation as introduced in Section~\ref{subs: Fine-tuning plus Transfer learning}. To derive a single label for each sentence pair, the median of the individual annotations is used. The Table~\ref{tab:Dataset Distribution} shows the distribution of the final processed data, which includes $11,353$ samples. And we randomly split the data into train/dev/test with 80\%/10\%/10\%.

\begin{table}
\footnotesize
\setlength{\tabcolsep}{5pt}
\begin{center}
\begin{tabular}{lcccccc}
\toprule
   naturalness & 1 & 2 & 3 & 4 & 5 & \color{red} 6    \\
   \midrule
   data size & 394 & 373 & 670 & 2,185 & 3,062 & \color{red}4,669   \\
   \midrule
   total & \multicolumn{6}{c}{11,353} \\
   \bottomrule
\end{tabular}
\end{center}
\caption{\label{tab:Dataset Distribution} Distribution of naturalness labels of the data set with the majority label marked red.}
\end{table}

\subsection{Fine-tuning BERT}
\label{subs: Fine-tuning BERT}

During fine-tuning, the entire pre-trained BERT model is optimised end-to-end. The output of token [CLS]: the pooled\_output, is further fed to a linear layer with softmax activation function with parameters $W \in {\rm I\!R}^{K \times H}$, where $H$ is the dimension of the hidden state vectors and $K$ is the number of classes. In this paper, we applied English uncased BERT-Base model~\footnote{https://tfhub.dev/google/bert\_uncased\_L-12\_H-768\_A-12/1}, which has 12 layers, 768 hidden states and 12 heads, for the naturalness classification. So $H$ is 768 and $K$ is 6 in our experiment. All hyper-parameters are tuned to our dataset. The batch size is 256 and the number of epochs is 25. Adam~\citep{kingma2014adam} is used for optimization to minimize the cross-entropy with an initial learning rate of 5e-3.

\subsection{Baselines}
\label{subs: baselines}

As this is a novel task, there is no available existing baseline for us to compare. Hence, we apply the following three baselines for performance comparison in order to show the robustness of our proposed method.

\paragraph{BoW + SVM:} We firstly introduce a SVM classifier using BoW representation as baseline. The SVM~\citep{suykens1999least} is a discriminative classifier formally defined by a separating hyperplane and is widely used for classification task because of significant accuracy with less computation power. The BoW model~\citep{zhang2010understanding} is a text representation that counts the occurrence of words within a document. The BoW approach is very simple and flexible, and can be used for extracting word features from documents. These numerical BoW vector features are used as input to a SVM with linear kernel having hyper-parameters $C=1.0$ and $\gamma$ set to 'auto'.

\paragraph{Bi-LSTM:} We also introduce the bidirectional LSTM (Bi-LSTM) with one layer for naturalness evaluation as baseline. The Bi-LSTM layer has the same hidden size as fine-tuning BERT, i.e., 768, and the output is forward to one linear layer with softmax function for naturalness classification. We almost remain the same hyper-parameters setting as BERT for Bi-LSTM training, i.e., 256 batch size, 25 epochs and 5e-3 initial learning rate. We restrict the number of LSTM-layers to one because multiple layers resulted in very slow training speed, as the LSTM cannot be trained in parallel, and worse performance.


\paragraph{BLEURT:} In order to establish robust performance comparison, we also apply a pre-trained model for naturalness evaluation as the third baseline: BLEURT~\citep{sellam2020bleurt}, which is a machine learned automatic metric for text generation. BLEURT continually pre-trained BERT~\citep{devlin2019bert} with a large number of synthetic reference-candidate pairs on several lexical- and semantic-level supervision tasks and then fine-tuned on multiple human ratings.~\citet{sellam2020bleurt} published multiple versions of BLEURT in the official repository\footnote{https://github.com/google-research/bleurt}, which includes BLEURT-tiny, BLEURT-base and BLEURT-large. More information could be found in the link\footnote{https://github.com/google-research/bleurt/blob/master/checkpoints.md}. \cite{sellam2020bleurt} demonstrated that BLEURT is much closer to human annotation and also recommended to fine-tune the pre-trained BLEURT for custom applications. Hence, we apply the BLEURT-tiny in this work for naturalness classification by adding one additional linear layer with softmax function. We also tried the recommended BLEURT-base model for naturalness evaluation, however, it directly resulted in worse performance in our case.

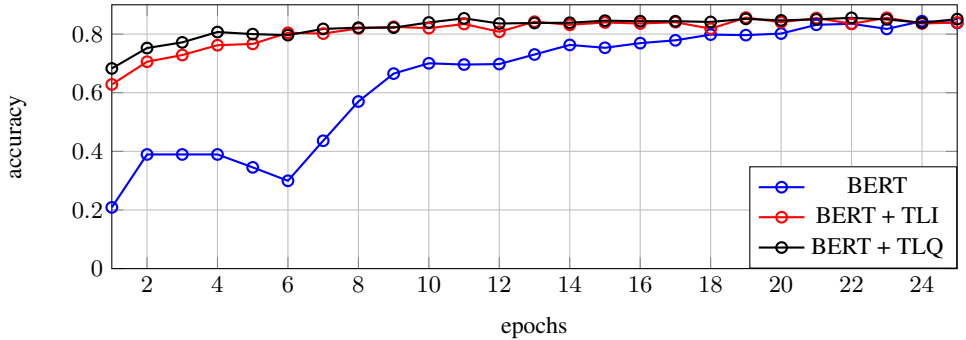
\begin{figure*}[!ht]
\centering
\begin{tikzpicture}[scale=1]
        \footnotesize
        \begin{axis}[
        legend style={at={(1,0)},anchor=south east,fill=white, fill opacity=1, draw opacity=1},
        ymin=0,
        ymax=0.9,
        width=5in,
        height=2in,
        grid=both,
        xmin = 1,
        xmax = 25,
        ylabel = accuracy,
        xlabel = epochs,
        ]
        
        \addplot[blue,thick,mark=o,mark options={solid}]
        table[x=a,y=b]{a b
        1 2.088105678558349609e-01
        2 3.894273042678833008e-01
        3 3.894273042678833008e-01
        4 3.894273042678833008e-01
        5 3.453744351863861084e-01
        6 2.995594739913940430e-01
        7 4.361233413219451904e-01
        8 5.700440406799316406e-01
        9 6.651982665061950684e-01
        10 7.004405260086059570e-01
        11 6.960352659225463867e-01
        12 6.977973580360412598e-01
        13 7.303964495658874512e-01
        14 7.629956007003784180e-01
        15 7.533039450645446777e-01
        16 7.691630125045776367e-01
        17 7.788546085357666016e-01
        18 7.982378602027893066e-01
        19 7.964757680892944336e-01
        20 8.017621040344238281e-01
        21 8.317180871963500977e-01
        22 8.352422714233398438e-01
        23 8.176211714744567871e-01
        24 8.440528512001037598e-01
        25 8.378854393959045410e-01
        };\addlegendentry{BERT}
        
        \addplot[red,thick,mark=o,mark options={solid}]
        table[x=a,y=b]{a b
        1 6.281938552856445312e-01
        2 7.057268619537353516e-01
        3 7.286343574523925781e-01
        4 7.621145248413085938e-01
        5 7.665198445320129395e-01
        6 8.044052720069885254e-01
        7 8.017621040344238281e-01
        8 8.193832635879516602e-01
        9 8.246695995330810547e-01
        10 8.202643394470214844e-01
        11 8.343612551689147949e-01
        12 8.079295158386230469e-01
        13 8.422907590866088867e-01
        14 8.317180871963500977e-01
        15 8.396475911140441895e-01
        16 8.361233472824096680e-01
        17 8.405286073684692383e-01
        18 8.185021877288818359e-01
        19 8.563876748085021973e-01
        20 8.405286073684692383e-01
        21 8.546255230903625488e-01
        22 8.352422714233398438e-01
        23 8.555065989494323730e-01
        24 8.361233472824096680e-01
        25 8.396475911140441895e-01
        };\addlegendentry{BERT + TLI}
        
        \addplot[black,thick,mark=o,mark options={solid}]
        table[x=a,y=b]{a b
        1 6.828193664550781250e-01
        2 7.524229288101196289e-01
        3 7.718061804771423340e-01
        4 8.070484399795532227e-01
        5 8.000000119209289551e-01
        6 7.964757680892944336e-01
        7 8.176211714744567871e-01
        8 8.220264315605163574e-01
        9 8.220264315605163574e-01
        10 8.396475911140441895e-01
        11 8.537445068359375000e-01
        12 8.361233472824096680e-01
        13 8.378854393959045410e-01
        14 8.387665152549743652e-01
        15 8.458150029182434082e-01
        16 8.440528512001037598e-01
        17 8.440528512001037598e-01
        18 8.414096832275390625e-01
        19 8.519823551177978516e-01
        20 8.466960191726684570e-01
        21 8.502202630043029785e-01
        22 8.555065989494323730e-01
        23 8.502202630043029785e-01
        24 8.387665152549743652e-01
        25 8.511013388633728027e-01
        };\addlegendentry{BERT + TLQ}
        
        \end{axis}

        \end{tikzpicture}
\caption{Comparison of the different approaches to naturalness estimation with respect to the training epochs. With additional transfer learning using the quality score (BERT + TLQ), the training speed is increased the most.}
\label{fig:comparing results}
\end{figure*}

\subsection{Fine-tuning BERT plus Transfer Learning}
\label{subs: Fine-tuning plus Transfer learning}

Analysing the correlation of the naturalness scores with the respective quality and informative scores using Spearman rank correlation coefficient~\citep{hauke2011comparison} shows the positive results. The spearman's correlation between naturalness and quality is $\rho = 0.60$ and between naturalness and informativeness is $\rho = 0.45$. Hence, this positive correlation is further leveraged through transfer learning~\citep{pan2009survey} using the same BERT-based setup. The procedure is as follows: first, the BERT model is fine-tuned to quality (or informativeness, respectively), and then this already fine-tuned model is continually fine-tuned once more using the naturalness score as target.

\section{Results and Discussion}
\label{sec: Results and Discussion}
The results of our proposed approach (BERT) for estimating the naturalness of a generated sentence given an additional reference are depicted in Table~\ref{tab: Performance Comparsion}. In addition to the baselines: BoW + SVM and Bi-LSTM and BLEURT, the majority class accuracy is also shown. It is calculated as the proportion of the majority class (naturalness score 6) resulting in $4669/11353=0.41$. BERT + TLI and BERT + TLQ represent the Transfer Learning results from Informativeness (TLI) and Quality (TLQ) respectively.

\begin{table*}
\footnotesize
\begin{center}
\begin{tabular}{lccccccc}
\toprule
    & majority class & BLEURT & BoW + SVM & Bi-LSTM & BERT &  BERT + TLI &  BERT + TLQ \\

   \midrule
   F1\_score & - & 0.13 & 0.66 & 0.69 & 0.83 & 0.84 & 0.86  \\
   recall & - & 0.22 & 0.66 & 0.66 & 0.83 & 0.83 & 0.84  \\
   precision & - & 0.18 & 0.67 & 0.73 & 0.82 & 0.85 & 0.89  \\
   accuracy & 0.41 & 0.42 & 0.68 & 0.74 & 0.85 & 0.87 & 0.88  \\
   \bottomrule
\end{tabular}
\end{center}
\caption{\label{tab: Performance Comparsion} Performance comparison of different methods shows the superiority of our proposed BERT on naturalness evaluation. }
\end{table*}

The results in Table~\ref{tab: Performance Comparsion} show that all BERT-based approaches outperform the baselines for classifying the naturalness of a generated sentence achieving a higher overall accuracy.

Even though the data set is imbalanced as shown in Table~\ref{tab:Dataset Distribution},~\citet{madabushi2019cost} indicate that BERT is capable of handling imbalanced classes with no additional data augmentation, which is also confirmed in our work. The Table~\ref{tab:Accuracy of different naturalness labels on BERT with test data} shows the accuracy of different naturalness score on our proposed model BERT with test data. Even if the data is seriously imbalanced, every naturalness class has comparative accuracy.

\begin{table}
\footnotesize
\setlength{\tabcolsep}{5pt}
\begin{center}
\begin{tabular}{lcccccc}
\toprule
   naturalness & 1 & 2 & 3 & 4 & 5 &  6    \\
   \midrule
   test size & 41 & 35 & 65 & 227 & 305 & 462   \\
   \midrule
   prediction size & 37 & 27 & 46 & 180 & 240 & 441   \\
   \midrule
   accuracy & 0.90 & 0.77 & 0.71 & 0.79 & 0.79 & 0.95   \\
   \bottomrule
\end{tabular}
\end{center}
\caption{\label{tab:Accuracy of different naturalness labels on BERT with test data} Accuracy of different naturalness scores on BERT with test data.}
\end{table}

Given the imbalanced data set, the macro F1\_score, recall and precision are also computed to show the robustness of our proposed approach. Moreover, through the transfer of quality (or informativeness) knowledge to naturalness training, the performance of naturalness estimation is further improved and training speed has also been greatly promoted. Figure~\ref{fig:comparing results} shows that the naturalness training based on transfer learning is faster and tends to be stable after only 5 epochs. Table~\ref{tab: Performance Comparsion} shows that transferring knowledge from quality results in the highest improvement on naturalness estimation. This is also consistent with the Spearman correlation of the naturalness scores and quality scores which is higher than the correlation of the naturalness scores and informativeness scores.

The Table~\ref{tab: Performance Comparsion} shows that using the BLEURT model for naturalness evaluation results in the worst performance even though BLEURT was already pre-trained on multiple tasks. The possible reason is that the BLEURT was pre-trained with multiple automatic metrics, hence, it has no superiority in our naturalness classification task.

\section{Conclusion and Future Work}
\label{sec: Conclusion and Future Work}

In this paper, we proposed a novel task of automatically estimating the naturalness for task-oriented generation based on a human reference. We proposed a robust estimation approach by fine-tuning the pre-trained BERT model which outperforms an SVM classifier, Bi-LSTM, fine-tuned BLEURT and majority class accuracy. Taking advantage of the positive correlation of naturalness on quality (or informativeness), we successfully improved the naturalness training speed and estimation performance through transfer learning.

This work sheds light on research towards naturalness evaluation by neural network learning. The final goal of our work is to relieve the human labors from naturalness evaluation task and realize the automatic naturalness evaluation for dialogue generation. Hence, we will firstly collect more human evaluation data for future work. Because the human evaluation data, which is already shared and public on the internet, is very limited. With more collected human evaluation data, we are also interested in the performance of our proposed method on other evaluation criteria, such as quality, coherence etc. And we will further verify the performance of our proposed method on chit-chat and open domain dialogue generation. 




\bibliographystyle{acl_natbib}
\bibliography{anthology,ranlp2021}


\end{document}